\pdfoutput=1

\documentclass[11pt]{article}

\usepackage[final]{acl}

\usepackage{times}
\usepackage{latexsym}

\usepackage[T1]{fontenc}

\usepackage[utf8]{inputenc}

\usepackage{microtype}

\usepackage{inconsolata}
\usepackage{listings}
\definecolor{backcolour}{rgb}{0.965,0.965,0.965}
\usepackage[T1]{fontenc}
\usepackage[scaled=0.85]{beramono}            
\lstdefinestyle{mystyle}{
    backgroundcolor=\color{backcolour},   
    basicstyle=\ttfamily\footnotesize\bfseries,
    breakatwhitespace=false,         
    breaklines=true,                 
    captionpos=t,                    
    keepspaces=true,                 
    numbersep=5pt,                  
    showspaces=false,                
    showstringspaces=false,
    showtabs=false,                  
    tabsize=2,
    frame=single,
    framerule=0.0mm,    
    framesep=5pt,
    xleftmargin=5pt,
    xrightmargin=5pt,
}
\usepackage{graphicx}
\usepackage{tabularx} 
\usepackage{multirow}
\usepackage{booktabs} 
\usepackage{diagbox}
\usepackage{amsmath}
\title{Adapting Abstract Meaning Representation Parsing to the Clinical Narrative -- the SPRING THYME parser}

\author{Jon Z. Cai\textsuperscript{1}, {\bf Kristin Wright-Bettner}\textsuperscript{1}\\ {\bf Martha Palmer}\textsuperscript{1}, {\bf Guergana K. Savova}\textsuperscript{2}, {\bf James H. Martin}\textsuperscript{1} \\
        \textsuperscript{1}University of Colorado Boulder \\ \textsuperscript{2}Boston Children’s Hospital and Harvard Medical School}

\begin{document}
\maketitle
\begin{abstract}
This paper is dedicated to the design and evaluation of the first AMR parser tailored for clinical notes. Our objective was to facilitate the precise transformation of the clinical notes into structured AMR expressions, thereby enhancing the interpretability and usability of clinical text data at scale. Leveraging the colon cancer dataset from the Temporal Histories of Your Medical Events (THYME) corpus, we adapted a state-of-the-art AMR parser utilizing continuous training. Our approach incorporates data augmentation techniques to enhance the accuracy of AMR structure predictions. Notably, through this learning strategy, our parser achieved an impressive F1 score of 88\% on the THYME corpus's colon cancer dataset. Moreover, our research delved into the efficacy of data required for domain adaptation within the realm of clinical notes, presenting domain adaptation data requirements for AMR parsing. This exploration not only underscores the parser's robust performance but also highlights its potential in facilitating a deeper understanding of clinical narratives through structured semantic representations.
\end{abstract}

\section{Introduction}
Abstract Meaning Representation \cite{banarescu-etal-2013-abstract}(AMR)is a highly adaptable and expressive framework designed to capture the semantics of natural language expressions. Automatic AMR parsing is a natural language processing (NLP) method that translates natural language inputs into formal AMR expressions – representations which have proven to be useful across a wide range of downstream applications \cite{kapanipathi-etal-2021-leveraging, liu-etal-2015-toward, liao-etal-2018-abstract, li-flanigan-2022-improving, bonial-etal-2020-dialogue, bai-etal-2021-semantic} including those in the biomedical domain \cite{garg2016, rao-etal-2017-biomedical}. 

Formally, AMR expressions take the form of labeled, rooted, directed, and acyclic graphs, $g=(V, E)$, where $V$ represents the set of AMR nodes, which can be of type predicate, abstract concept and attributes; $E$ represents the possible semantic relations between nodes such as prototypical agent and patient denoted by \texttt{arg0} and \texttt{arg1}. The AMR graph structure underpinned by Neo-Davidsonian semantics can then effectively encapsulate the abstract concepts, relationships, and entities present in individual sentences or utterances. 

From a practical standpoint, AMR expressions encompass the semantic content typically addressed by individual representation schemes such as semantic role labeling \cite{palmer-etal-2005-proposition}, named entities \cite{NNER}, and coreference chains \cite{joshi-etal-2020-spanbert}, thereby unifying these diverse aspects of meaning into a single comprehensive representation. Figure 1 illustrates an AMR expression selected from the clinical domain. 
\begin{figure}
    \centering
    \includegraphics[width=\columnwidth]{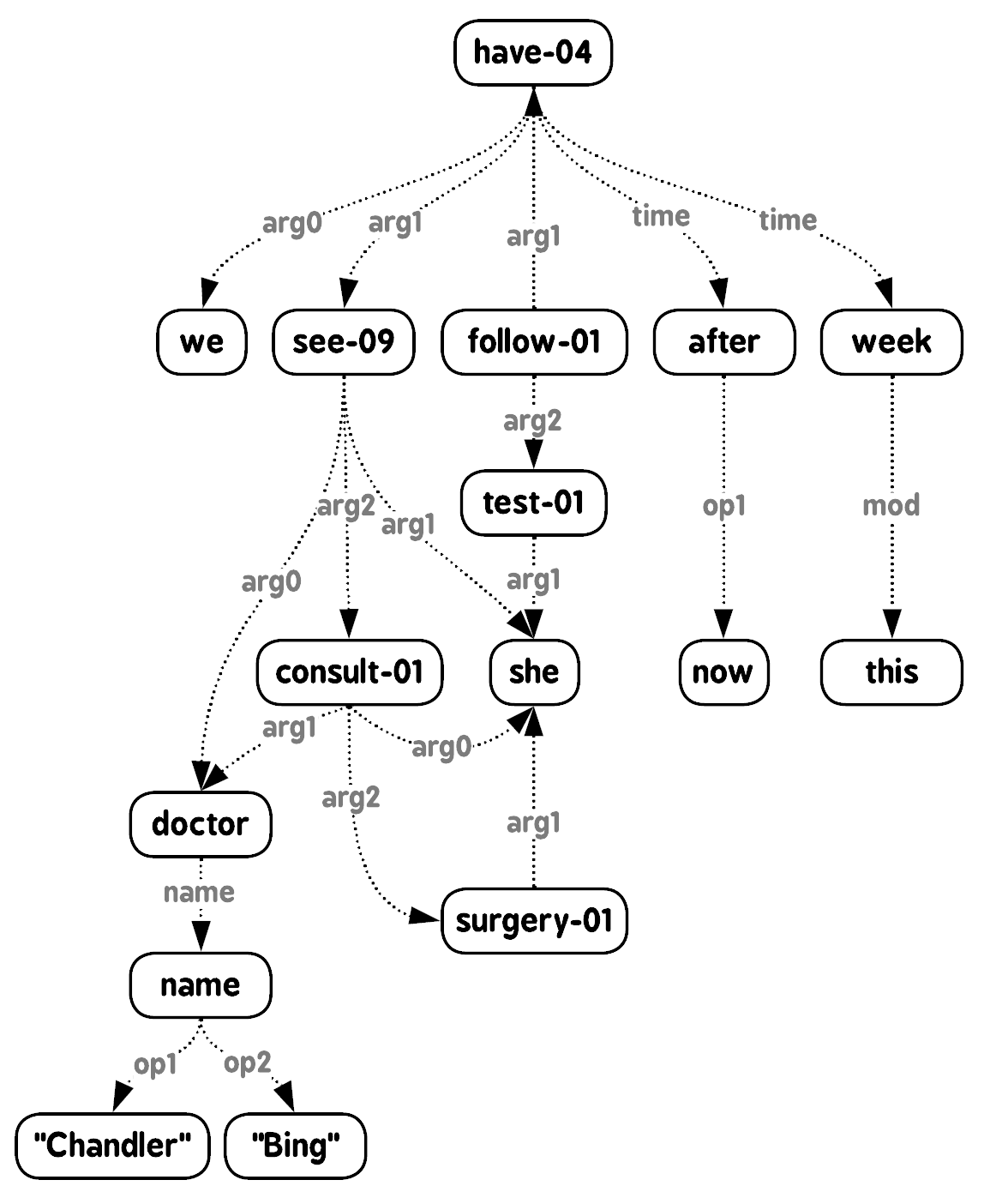}
    \caption{the AMR graph of sentence ``We will have her see Dr. Chandler Bing in surgical consultation later this week following her testing.''}
    \label{fig:amr-example}
\end{figure}

As Figure \ref{fig:amr-example} demonstrates, concepts including events, entities and properties are captured as nodes in the graph, while the relations among the concepts are captured by labeled edges connecting the nodes. Events are represented using PropBank frames \cite{palmer-etal-2005-proposition}, and the semantic relations of both entities and events to these predicates are specified either by a frame’s numbered argument or one of the relations from AMR’s role inventory. For example, the see-09 predicate represents the event of ``visit/consultation by a medical professional.'' In this case, the agent of the seeing event is ``Dr. Chandler Bing'', represented by \texttt{see-09}’s \texttt{ARG0} relation, and the semantic role of patient for the event is ``she'' indicated by the \texttt{ARG1} semantic relation. AMR graphs also specify the temporal information in a formal way. In the above example, the time of the seeing event is specified by two temporal modifier subgraphs. It is a conjunction of ``after now'' and ``within this week'' which makes ``later this week'' a concrete time range.

AMR parsers based on pretrained large language models and sequence-to-sequence (encoder-decoder) architectures have demonstrated impressive accuracy when trained and evaluated on standard datasets. The use of AMR parsers has contributed to improved performance across a range of NLP tasks including question answering \cite{fu-etal-2021-decomposing-complex}, information retrieval \cite{liao-etal-2018-abstract}, knowledge-graph construction \cite{ribeiro-etal-2022-factgraph}, and text generation \cite{bai-etal-2022-graph}. 

These successes have sparked growing interest in employing AMR in domains that diverge from the existing training data, such as human-robot interaction tasks, educational applications involving classroom discourse analysis, and diverse biomedical use cases. Unfortunately, as language form and meaning deviate from the general language captured in generic training data, parsing performance shows a rapid decline. This decline stems from disparities in vocabulary, syntax, and overall discourse structure. Addressing these challenges necessitates dedicated human expert annotation efforts to create domain-specific AMR resources. However, such endeavors can be costly and time-consuming. Hence, the preference lies in maximizing the utilization of existing data and parsers and adapting them to new domains, rather than building entirely new systems from scratch.

The contributions of this paper include:
\begin{itemize}
    \item We adapted the high-performance SPRING parser \cite{bevilacqua-etal-2021-one} to the clinical domain, specifically leveraging the Temporal Histories of Your Medical Events (THYME) corpus \cite{wright-bettner-etal-2020-defining}, and achieved state-of-the-art performance in AMR parsing within this context..
    \item We demonstrated that by tailoring an existing general domain English neural AMR parser with a relatively modest amount of gold-standard in-domain data, we could attain significantly high accuracy.
    \item We showcased data augmentation techniques that effectively enhance the parser's robustness across different domains.
\end{itemize}

\section{Data}
Supervised training data for AMR parsers consists of pairs of linguistic expressions along with their associated human annotated gold-standard AMR expressions. The current standard dataset for AMR development is AMR 3.0 \cite{amr-annotation-release3.0} available from the Linguistic Data Consortium as LDC2020T02.  This general domain dataset is the basis for our baseline efforts prior to domain adaptation.  AMR 3.0 consists of over 59k English expressions from a variety of broadcast conversations, newswire, weblogs, web discussion forums, fiction and web text. To  facilitate evaluation and model comparison, AMR 3.0 is divided into standard training, development and test splits consisting of 55,635, 1,722, and 1,898 expressions respectively. 

To adapt AMR to the clinical narrative, we developed 8,327 in-domain AMRs (separate paper with detailed description under review) on a subset of the THYME colon cancer corpus \cite{styler2014, wright-bettner-etal-2020-defining}. The colon cancer part of the THYME corpus consists of 594 de-identified physicians’ notes for 198 patients with colon cancer. Each patient is represented by one pathology note and two clinical notes. The corpus has undergone several prior annotation efforts, including temporal and coreference annotation \cite{styler2014, wright-bettner-etal-2019-cross, wright-bettner-etal-2020-defining} and entity tagging as defined by the Unified Medical Language System (UMLS \cite{journals/nar/Bodenreider04}).
As part of our AMR annotation process, we adopted seven clinical-domain named entity (NE) types (anatomical-site, clinical-attribute, devices, disease-disorder, medications-drugs, sign-symptom) from the UMLS project and relied heavily on the UMLS in classifying many AMR concepts. 

Like other genre-specific AMR tasks \cite{bonial-etal-2019-abstract, bonn-etal-2020-spatial}, we found it necessary to modify the standard AMR annotation approach to support meaningful annotation of domain-unique linguistic phenomena. Two phenomena are pervasive in the clinical narrative. First, physician notes  frequently drop eventive mentions when they are inferable by human readers. For example, ``Declines tetanus'' does not mean the patient declined having tetanus; they declined a tetanus immunization. We expanded AMR's guidelines to permit explicit rendering of certain implicit concepts like the immunization:

\begin{lstlisting}
(d / decline-02
      :ARG1 (s / shot-13 :implicit +
            :ARG3 (d2 / disease-disorder :name (n / name :op1 "tetanus"))))    
\end{lstlisting}

Second, like other specialized domains, clinical texts are rife with semantically dense noun phrases (NPs)  \cite{gron-etal-2018-interplay}. In AMR, NPs must be treated in one of two ways: Either all components are extracted and related (white marble = marble that is white), or they are analyzed as single units of meaning, i.e., NEs (White House). However, semantic compositionality exists on a spectrum \cite{Nakov2013OnTI}, and many specialized NPs in particular strain the adequacy of a binary approach. This can be seen even in simple clinical NPs: One annotator might decide ``blood pressure'' is a single, cohesive unit of meaning and annotate it as an NE, while another might decide ``pressure'' is an extractable property of ``blood''. To address this, we implemented a two-pass strategy: In the first pass, for NPs that fell under one of the clinical NE types mentioned above, an experienced annotator made these compositionality judgments and added each unique phrase to a searchable, phrasal NE Dictionary along with an AMR fragment that “defined” the compositionality for each phrase. Annotators then referenced the Dictionary when building the AMR graphs in the second pass. This approach supported consistency and speed of annotation.

Finally, the THYME corpus contains frequent repetition of many other multiword expressions and phrases. For extremely formulaic phrases, such as those found in Vital Signs sections (Height = 167.60 cm, e.g.), we implemented a template-filling script that deterministically produced the AMRs, again saving significant manual annotation time. Of the 8,327 AMRs, 1,640 were produced by this script; the rest were created manually. The final 8,327 THYME-AMR data are split into training, development and test sets randomly with 4,955, 1,641 and 1,731 sentence-AMR pairs, respectively. All of the model training is conducted on the training set of the AMR 3.0 and THYME AMR corpora. We show the Inter Annotator Agreement between three annotators on 107 THYME-AMRs in Table \ref{tab:IAA}

\begin{table}[h]
\resizebox{\columnwidth}{!}{%
\begin{tabular}{|c|c|c|c|}
\hline
Comparison    & P & R & F1 \\ \hline
gold vs annotator 1 &   0.93   & 0.93  & 0.93 \\ 
gold vs annotator 2 &  0.93 & 0.93 & 0.93 \\ 
annotator 1 vs annotator 2 & 0.91 & 0.90 & 0.90 \\ \hline
\end{tabular}%
}
\caption{Smatch scores on 107 manuall THYME AMRs, representing three clinical notes\label{tab:IAA}}%
\end{table}

\section{Methods}
We treat the AMR parsing task as a supervised machine learning problem and train a parameterized model to map natural language expressions to their corresponding AMR graphs. Various model architectures and training methods and paradigms have been employed over the years \cite{flanigan-etal-2014-discriminative, foland-martin-2017-abstract, lyu-titov-2018-amr, cai-lam-2019-core, zhang-etal-2019-amr, wang-etal-2015-transition, ballesteros-al-onaizan-2017-amr, fernandez-astudillo-etal-2020-transition, hoang2021ensembling}, resulting in a continuous improvement in the state of the art on the general domain AMR dataset(i.e. AMR 2.0 and 3.0 corpus (LDC2020T2)). However, these improvements are highly dependent on the availability of significant amounts of annotated training data hampering the development of parsers for specific genres and languages other than English.  Our approach here is to leverage an existing high-performance parser and adapt it to the clinical domain using the modest amount of domain-specific training data described in the last section.

Meanwhile, the great advances of the pre-trained foundational models has introduced a new modeling paradigm in the field of NLP as well as to structure-prediction problems such as AMR parsing. In particular, the sequence-to-sequence modeling, originally developed for machine translation, has proven a highly effective approach for AMR parsing \cite{bevilacqua-etal-2021-one, konstas-etal-2017-neural, xu-etal-2020-improving}.  In this approach, two neural network components are involved: an encoder, which takes the natural language sentence as input and maps it to a continuous manifold as a sequence of high-dimensional vectors, and a decoder, which takes the embedded sentence representation vectors and maps them to the output embedding space, corresponding to the target sequence tokens. 

\begin{figure}
    \centering
    \includegraphics[width=0.8\columnwidth]{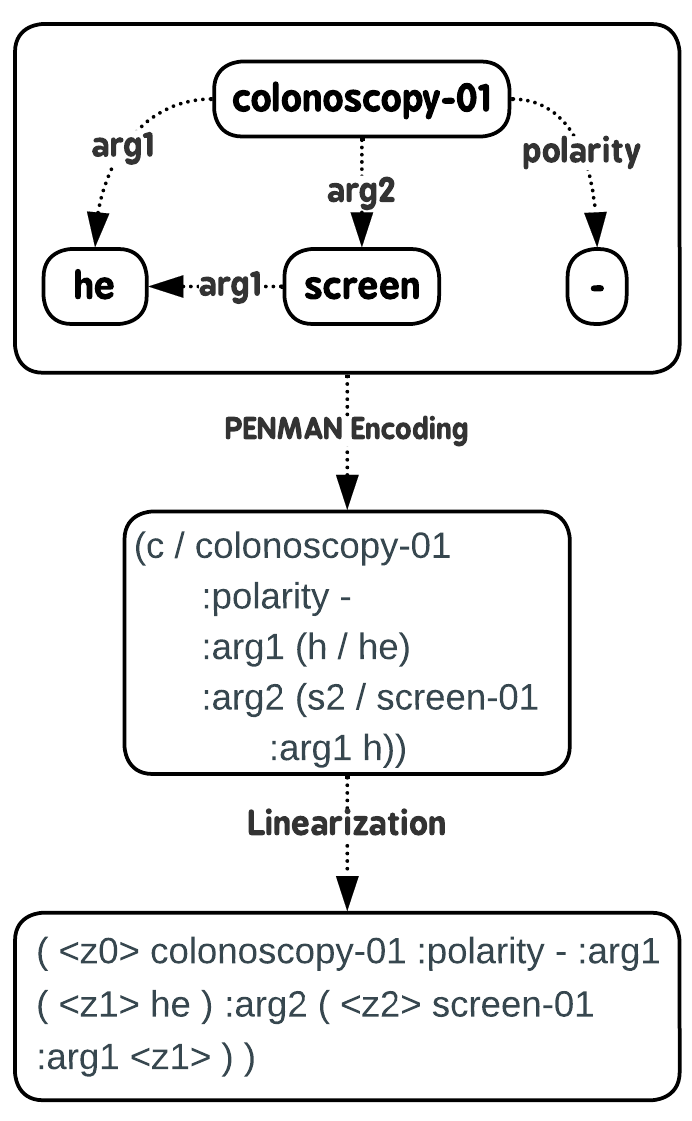}
    \caption{AMR graph to PENMAN linearization pipeline. The transformation map between the AMR graphical representation and its linearized representation is one-to-one-and-onto. }
    \label{fig:linearization}
\end{figure}

Here we make use of the SPRING parser \cite{bevilacqua-etal-2021-one}, one of the state-of-the-art AMR parsers on AMR 3.0 evaluation. The underlying pre-trained language model is BART-large \cite{lewis-etal-2020-bart}, a transformer-based language model that has been trained using a set of denoising pre-training objectives, such as a masked language modeling objective and a document reconstruction objective, on general domain unlabeled English text.  The neural network architecture relies on the self-attention and cross-attention mechanism to learn patterns from natural language texts. This pre-trained model is then fine-tuned on the AMR 3.0 training data to map English inputs to linearized AMR graphs, which consist of a sequence of AMR tokens. We show the linearization correspondence of an AMR graph to its sequence of AMR tokens in Figure \ref{fig:linearization}.

\begin{figure}
    \centering
    \includegraphics[width=\columnwidth]{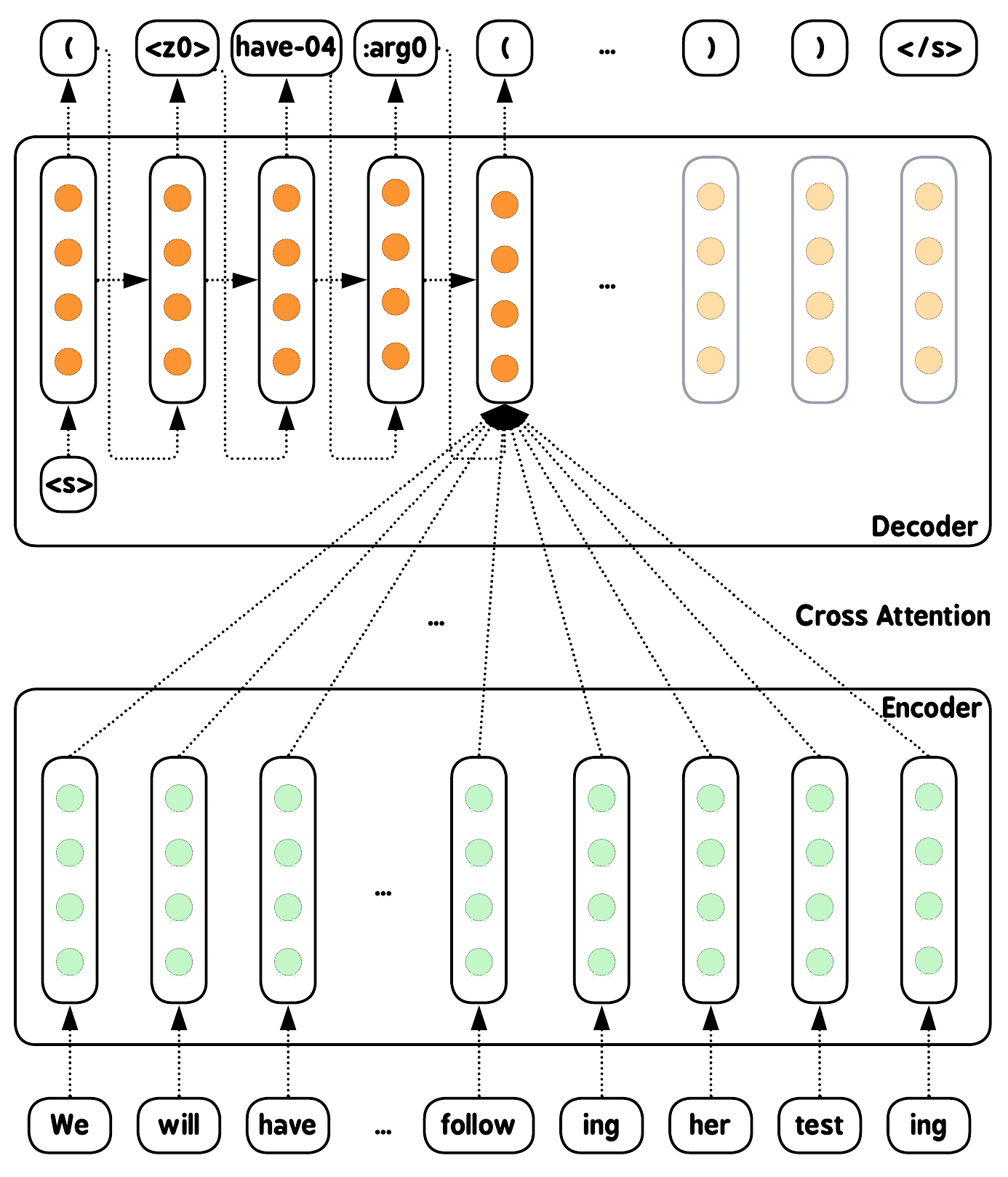}
    \caption{The SPRING parser modeling diagram. A transformer-based self-attention mechanism is used to produce embeddings for the input expression. The decoder then uses cross attention to drive autoregressive generation of a sequence of AMR output tokens.}
    \label{fig:encder-decoder}
\end{figure}

A critical aspect of using sequence-to-sequence models for structured prediction tasks, like parsing, is transforming the task itself. In AMR parsing, the AMR graph is converted into a sequence of tokens through a linearization algorithm. Note that the vocabulary of the decoder differs from that of the encoder model, as the target sequence consists of  AMR-specific tokens such as the relations \texttt{arg0} and \texttt{arg1}, and predicates like \texttt{test-01}. During  fine-tuning, we utilize the vocabulary derived from the AMR 3.0 corpus, which ensures consistency and accuracy in the parsing process. The parsing problem is then to convert an input text sequence into a valid sequence of AMR tokens that can be deterministically transformed into a directed AMR graph. The overall SPRING approach is depicted in Figure \ref{fig:encder-decoder}.
Given a high-performing SPRING model, we adapt it to the THYME domain by fine-tuning on the THYME-AMR training set (4,955 expressions).  Here, fine-tuning involves continuous gradient-based updates to the original model parameters with a small learning rate ($5\times 10^{-6}$) with batch size to be 20, we keep the maximum sequence length to be 1024..

\subsection{Evaluation}
The standard metric to evaluate AMR parsing performance is SMATCH, which decomposes an AMR graph into triples that capture the edge list representation of a graph structure. For instance, the AMR for the sentence ``He had never undergone a screening colonoscopy.'' can be decomposed into its edge list representation as AMR1 and edge list 1 as follows:

\begin{lstlisting}
AMR1: 
(c / colonoscopy-01 :polarity -
      :arg1 (h / he)
      :arg2 (s2 / screen-01
            :arg1 h))
            
AMR2: 
(c1 / colonoscopy-01 :polarity -
      :arg1 (s / she)
      :arg2 (s2 / screen-01
            :arg1 s))
\end{lstlisting}

\begin{lstlisting}
Decomposed edge list1:
      instance(c, colonoscopy-01)
      instance(h, he)
      instance(s2, screen-01)
      polarity(c, -)
      arg1(c, h)
      arg2(c, s2)
      arg1(s2, h)

Decomposed edge list2:
      instance(c1, colonoscopy-01)
      instance(s, she)
      instance(s2, screen-01)
      polarity(c1, -)
      arg1(c1, s)
      arg2(c1, s2)
      arg1(s2, s)
\end{lstlisting}

We conjured another slightly altered AMR2 with the \texttt{he} node replaced with a \texttt{she} node, indicating a potential mistake in the parser generated AMR. In the above decomposition of AMR graphs, \texttt{instance()} represents the nodes in the graph while the rest are the edges. Given the edge lists for a hypothetical parse and its corresponding gold-standard parse, the SMATCH metric produces precision (p), recall (r), and F1-measure scores as follows:
\begin{align*}
    p = \frac{N_{\textit{correct}}}{N_{\textit{predicted}}}, r = \frac{N_{\textit{correct}}}{N_{\textit{reference}}}, F_1 = \frac{2p r}{p+r}
\end{align*}

A complication in computing these scores is that we need to know which of the proposed AMR nodes in the parse are supposed to correspond to which ones in the correct set.  In other words, the graphs need to be matched before they can be scored. This issue originates from the encoding of AMR nodes with variables, through which different instantiations of a concept can be encoded. The standard SMATCH scorer \cite{cai-knight-2013-smatch} employs a greedy heuristic method to provide the required alignment to avoid  computing a computationally expensive optimal alignment.

Finally, AMR representations are an amalgamation of semantic representations including predicate-argument relations, named entities, and coreference components.  The SMATCH score represents an average over these component categories, obscuring the model performance over the various categories of information in AMR expressions, thus making it difficult to assess the usability of the results in downstream applications. To address this, a more fine-grained analysis tool\footnote{\url{https://github.com/mdtux89/amr-evaluation}} provides precision, recall and F1 measures across the various component AMR tasks. We will discuss the fine-grained categories in section \ref{sec:fine-grained-performance}.

\section{Experiments}
We present the domain adaption training experiments in this section to show the characteristics of the text from THYME corpus when it comes to AMR parser developement.
\subsection{Domain Adaptation}
Table \ref{tab1} provides the results of our primary domain adaptation experiments. The first column presents the evaluation results of the off-the-shelf SPRING AMR parser trained solely with the AMR 3.0  training data. The 83.0 SMATCH score for the SPRING parser reaches near state-of-the-art performance on the AMR 3.0 test set, whereas, the performance on the THYME-AMR test set is significantly lower at 51.7 SMATCH. The second column shows the results of the same parser fine-tuned using the THYME-AMR training data. Here, we see that the fine-tuned parser achieves excellent results on the THYME-AMR corpus test set with a 35.3 point absolute improvement over the original model.
\begin{table}[h]
\resizebox{\columnwidth}{!}{%
\begin{tabular}{|c|c|c|c|}
\hline
\diagbox{Test}{Train}    & AMR 3.0 & THYME-AMR & \begin{tabular}[c]{@{}c@{}}AMR 3.0 \\ + THYME-AMR\end{tabular} \\ \hline
AMR 3.0 &   83.0   & 77.0  & 80.0 \\ \hline
THYME-AMR     &  51.7 & 87.0 & 88.0 \\ \hline
\end{tabular}%
}
\caption{SPRING THYME-AMR parser performance with different training sources. All scores are Smatch F1\label{tab1}}%
\end{table}
\subsection{Avoiding Forgetting}
Catastrophic forgetting is a frequently observed problem when fine-tuning large pre-trained models on domain specific data \cite{li-hoiem, riemer2018learning, scialom-etal-2022-fine}. While fitting the model’s parameters to the new domain, there is often a significant loss in terms of the model’s performance on its original domain.  To assess the robustness and potential forgetting of general domain AMR knowledge, we evaluated the THYME-AMR fine-tuned parser on the AMR 3.0. The results showed a decrease in performance from 83.8 to 77.0, indicating significant forgetting of the general domain AMR.

Based on this observation, we deployed a joint training approach to mitigate this forgetting phenomenon. In this experiment, we fine-tuned the parser on a mixture sampled from both the AMR 3.0 and THYME-AMR data. Considering the differing sizes of the two corpora, we sampled them in a 12-to-1 ratio between THYME-AMR and AMR 3.0 sources. As can be seen from Table \ref{tab1}, this modest infusion of general domain data allowed the parser to attain high performance on the THYME-AMR test set while also largely maintaining its performance on the AMR 3.0 test set.
This observation underscores the effectiveness of domain-specific annotation in improving semantic parsing in a joint fashion. This means that the understanding of semantics improves collectively rather than independently, thanks to domain-specific data. As more representative data are collected, we expect further improvements in the parser's performance, making it even more adept at comprehending the semantics in the given domain.

\subsection{Fine-Grained Performance}\label{sec:fine-grained-performance}
Table \ref{tab2} presents detailed results of our best-performing parser across the semantic components that comprise AMR graphs. AMR representations are an amalgamation of semantic representations including predicate-argument relations, named entities, and coreference components.  The SMATCH score represents an average over these sub-categories. To leverage the in-depth analytical power of these linguistic sub categories, a more fine-grained analysis tool\footnote{\url{https://github.com/mdtux89/amr-evaluation}} provides precision, recall and F1 measures across the various component AMR tasks. We list the fine-grained performance metric category definitions briefly as follows: \\
\begin{itemize}
    \item \emph{Unlabeled} category assesses the parsing performance on the AMR graph, disregarding the edge labels.
    \item \emph{No WSD} category evaluates the parsing performance while ignoring the Propbank word sense labels (e.g., \texttt{see-09} becomes just \texttt{see}).
    \item \emph{Concepts} category considers only the abstract concept node matches.
    \item \emph{Named Entity} category focuses on the matches of named entity subgraphs.
    \item \emph{Negation} category concerns the matches of the negation attribute nodes(e.g. the \texttt{:polarity} edges).
    \item \emph{Reentrancy} category examines only the concept re-entrancy subgraphs(usually a back reference node).
    \item \emph{Semantic Role Label (SRL)} category pertains to the performance of each predicate argument structure generation.
\end{itemize}

\begin{table*}[t]
\makebox[\textwidth]{%
\begin{tabular}{|c|c|c|c|c|}
\hline
Sub-category                  & Training Set & \multicolumn{1}{c|}{Precision} & \multicolumn{1}{c|}{Recall} & \multicolumn{1}{c|}{F1} \\ \hline
\multirow{3}{*}{SMATCH}       & THYME-AMR + AMR 3.0  & 0.89                           & 0.88                        & 0.88                    \\ \cline{2-5} 
                              & THYME-AMR        & 0.88                           & 0.87                        & 0.87                    \\ \cline{2-5} 
                              & AMR 3.0          & 0.53                           & 0.45                        & 0.49                    \\ \hline
\multirow{3}{*}{Unlabeled}    & THYME-AMR + AMR 3.0  & 0.90                           & 0.90                        & 0.90                    \\ \cline{2-5} 
                              & THYME-AMR        & 0.90                           & 0.88                        & 0.89                    \\ \cline{2-5} 
                              & AMR 3.0          & 0.60                           & 0.51                        & 0.55                    \\ \hline
\multirow{3}{*}{No WSD}       & THYME-AMR + AMR 3.0  & 0.89                           & 0.88                        & 0.88                    \\ \cline{2-5} 
                              & THYME-AMR        & 0.88                           & 0.87                        & 0.87                    \\ \cline{2-5} 
                              & AMR 3.0          & 0.55                           & 0.46                        & 0.50                    \\ \hline
\multirow{3}{*}{Concepts}     & THYME-AMR + AMR 3.0  & 0.93                           & 0.92                        & 0.93                    \\ \cline{2-5} 
                              & THYME-AMR        & 0.93                           & 0.91                        & 0.92                    \\ \cline{2-5} 
                              & AMR 3.0          & 0.52                           & 0.46                        & 0.49                    \\ \hline
\multirow{3}{*}{Named Ent.}   & THYME-AMR + AMR 3.0  & 0.94                           & 0.93                        & 0.93                    \\ \cline{2-5} 
                              & THYME-AMR        & 0.93                           & 0.92                        & 0.92                    \\ \cline{2-5} 
                              & AMR 3.0          & 0.18                           & 0.05                        & 0.08                    \\ \hline
\multirow{3}{*}{Negation}     & THYME-AMR + AMR 3.0  & 0.86                           & 0.85                        & 0.85                    \\ \cline{2-5} 
                              & THYME-AMR        & 0.84                           & 0.86                        & 0.85                    \\ \cline{2-5} 
                              & AMR 3.0          & 0.45                           & 0.42                        & 0.44                    \\ \hline
\multirow{3}{*}{Reentrancies} & THYME-AMR + AMR 3.0  & 0.78                           & 0.79                        & 0.78                    \\ \cline{2-5} 
                              & THYME-AMR        & 0.78                           & 0.76                        & 0.77                    \\ \cline{2-5} 
                              & AMR 3.0          & 0.48                           & 0.37                        & 0.41                    \\ \hline
\multirow{3}{*}{SRL}          & THYME-AMR + AMR 3.0  & 0.88                           & 0.87                        & 0.87                    \\ \cline{2-5} 
                              & THYME-AMR        & 0.87                           & 0.85                        & 0.86                    \\ \cline{2-5} 
                              & AMR 3.0          & 0.55                           & 0.47                        & 0.51                    \\ \hline
\end{tabular}%
}
\caption{SPRING parser performance analytical breakdowns comparison among three models trained on different combination of the fine-tuning data source. The evaluation is on the THYME-AMR test set.
\label{tab2}}%
\end{table*}
We observe that the mixed data augmentation technique significantly improves performance across the board, impacting almost every sub-category of evaluation. Notably, the off-the-shelf parser faced significant challenges in understanding the semantics in the new domain. The performance drop due to domain shifting was not uniform across different sub-categories. The most significant drop in performance was seen in \emph{Named Entity} Recognition, which is expected due to the abundance of medical-related terminology. On the other hand, the data-augmented parser excelled in \emph{Concept} predication and \emph{Named Entity} recognition aspects of AMR parsing, while the performance in the \emph{Negation} and \emph{Reentrancy} category was relatively less impressive compared to the other categories. 

\subsection{Data Requirements for Successful Adaptation}
\begin{figure}
    \centering
    \includegraphics[width=\columnwidth]{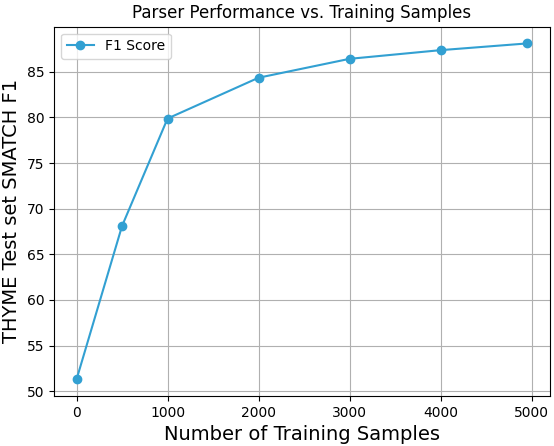}
    \caption{The performance curve with different sample sizes of the THYME-AMR training set. The x axis is the sample size of the training data; the y axis represents the SMATCH F1 performance score(with the unit of percentage) of the parsers evaluated on the same withheld test set (THYME-AMR test set)}
    \label{fig:parser-performance}
\end{figure}

Manual annotation of AMR data is time consuming and expensive. At the current time, the standard AMR 3.0 still consists of only 60k sentences, nearly 10 years after the initial data release.  The results shown in Table \ref{tab1} raise the question of how much data is actually required to attain high levels of parser accuracy through adaptation. To address this question, we conducted a series of experiments training models with progressively larger snapshots of the available training data. Specifically, we gradually augmented the training set size for each model by random sampling without replacement from the training data (resulting in training sets of size 500, 1,000, 2,000, 3,000, 4,000 and 4,955). The results in Figure \ref{fig:parser-performance} illustrate the parser's performance across these training sets. 

As can be seen, performance rapidly rises from the non-adapted baseline to 80 SMATCH with 1,000 training examples; the model trained on only 2,000 samples achieves 90\% of the performance of our best parser trained on all available training data. This rapid improvement with domain specific data  is a positive indication of the effectiveness of continued training from a generic model and its ability to rapidly generalize from the domain-specific data.
\section{Discussion}
Our results have highlighted the advantages of employing data augmentation techniques for domain adaptation fine-tuning. This opens up the possibility for additional follow-up studies, including the incorporation of data from domain-specific Propbank roleset development. For instance, in the case of THYME, leveraging example sentences for newly added named-entity types like ``anatomical-site'' could prove beneficial. Initializing the word embedding vectors with such domain-specific concepts would enable a better fit with the pre-trained foundational models. Future investigations involving more sophisticated foundational models and data augmentation approaches hold great promise for enhancing AMR parsing in the medical domain and other specialized domains. By harnessing the capabilities of cutting-edge language models and innovative data augmentation strategies, we can expect significant advancements in semantic parsing tasks and domain adaptation techniques.

With these advances, AMR parses have wide applicability to core information extraction tasks from the clinical narrative such as entity recognition, negation detection, uncertainty detection, coreference, temporality and relation extraction.

\section{Conclusion}
In our investigation, we have presented substantial evidence highlighting the critical role of domain-specific AMR annotations in the context of domain adaptation. Our findings illuminate how variances in the distribution between original and target domains can precipitate a marked decline in the performance of AMR parsing. This phenomenon underscores the challenge of catastrophic forgetting, a significant hurdle in the training of neural network models where new learning can disrupt previously acquired knowledge.

To counteract this issue, we demonstrated the critical role of data augmentation techniques. Specifically, by integrating domain-specific examples into the training dataset, we significantly bolstered the model's capability to acclimate to the nuances of the new domain while preserving its proficiency in the original domain. This strategic approach of coupling domain-specific annotation with thoughtful data augmentation has emerged as a formidable solution, ensuring both the robustness and accuracy of AMR parsing across different domain adaptation scenarios.

Our study reaffirms the indispensability of domain-specific annotation in achieving effective domain adaptation and also supports data augmentation as an essential tool in maintaining a delicate balance between learning new domain characteristics and retaining essential knowledge from the original domain. This balanced approach provides a promising avenue for future research and development in the field of AMR parsing, potentially paving the way for more nuanced and adaptable AI systems capable of navigating other domains with limited data yet maintain robustness.

\section{Limitations and Future Work}
Our study faced constraints primarily due to computational limitations, which necessitated a focus on a specific subset of model and data augmentation strategies. A reasonable extension of this research could involve the exploration of more advanced foundational models, including GPT-3.5, GPT-4, and their publicly accessible counterparts such as LLAMA. These platforms present opportunities for experimenting with zero- or few-shot learning techniques. Importantly, our use of clinical data mandates adherence to stringent privacy standards; thus, it is imperative that any models employed can be locally installed and operated within a secure, firewall-protected environment. This requirement currently excludes the use of proprietary models like those within the GPT family, which are tailored for commercial applications and do not meet the privacy criteria essential for our research objectives.

\section{Acknowledgements}
This research and the paper it informs have been funded by the United States National Institutes of Health (grants R01LM010090, R01LM013486). The content is solely the responsibility of the authors and does not necessarily represent the official views of the United States National Institutes of Health. The results presented in this paper were obtained using the Chameleon testbed\cite{keahey2020lessons} supported by the National Science Foundation. We wish to extend our heartfelt gratitude to Ahmed Elsayed and Gaby Dinh for their exceptional annotation efforts, without which this research would not have been possible. Additionally, we are deeply appreciative of the thorough discussions and invaluable suggestions from Skatje Myers, Steven Bethard, David Harris, Timothy Miller, Danielle Bitterman, Piet de Groen, and Dmitriy Dligach.

\section*{Ethics Statement}
In our exploration of clinical notes analysis and the design of automation systems, we navigate through a terrain rich with sensitive personal data and entwined with ethical complexities. Our work is fundamentally rooted in a profound respect for the dignity, rights, and welfare of the individuals whose lives and experiences are documented in these notes. Guided by a set of core ethical principles, our research endeavors to uphold the highest standards of integrity and respect.

Foremost, we prioritize the privacy and confidentiality of patient data. In this paper, all examples have been rigorously de-identified to ensure no personal information can lead back to individuals. Moreover, recognizing the critical importance of obtaining informed consent, we actively collaborate with institutional review boards (IRB) to ethically justify and secure consent approvals for utilizing all data involved in our research.

We are acutely aware of the potential biases in our analysis and interpretation of clinical narratives. This awareness extends to biases that might emerge from the data collection process, the selection of narratives for analysis, and our own preconceptions. We are committed to making concerted efforts to ensure that our analysis encompasses diverse perspectives, thereby avoiding the perpetuation of stereotypes or inequalities.

We urge downstream users of our parser to conscientiously consider the potential impact of their findings on the individuals depicted in the clinical narratives, as well as on wider patient populations. This involves thoughtful reflection on how the research could affect public perceptions, clinical practice, and policy making. A crucial aspect of our approach is to balance the dissemination of research findings with the imperative to prevent harm or distress.

Lastly, our pursuit of transparency in our methodology and findings is relentless. We advocate for the use of Abstract Meaning Representation (AMR) as a superior tool compared to opaque, ``black-box'' models. AMR offers a fully transparent and verifiable representation of the semantics in clinical narratives, which aligns with our commitment to fostering trust and accountability.

Our approach is a testament to our dedication to ethical research practices, emphasizing the protection of privacy, the mitigation of bias, the thoughtful consideration of impacts, and the advancement of transparency and accountability. These principles are the bedrock of our efforts to contribute meaningful and ethically sound advancements in the field of clinical notes analysis and automation system design.

\appendix

\end{document}